\documentclass[conference]{IEEEtran}
\IEEEoverridecommandlockouts
\usepackage{amsmath,amssymb,amsfonts}

\usepackage{tikz}
\usetikzlibrary{shapes,positioning,fit,decorations.pathreplacing}

\usepackage{amssymb}
\usepackage{pifont}
\newcommand{\cmark}{\ding{51}}%
\newcommand{\xmark}{\ding{55}}%

\usepackage{tikz}
\usetikzlibrary{shapes,arrows,positioning}

\usepackage{tabularx, booktabs, array}
\usepackage{multirow}

\usepackage[backend=biber,
            style=ieee,
            sorting=none,   
            maxnames=2    
]{biblatex}
\usepackage{algorithmic}
\usepackage{graphicx}
\usepackage{textcomp}
\usepackage{siunitx}
\usepackage{float}
\usepackage{xcolor}
\def\BibTeX{{\rm B\kern-.05em{\sc i\kern-.025em b}\kern-.08em
    T\kern-.1667em\lower.7ex\hbox{E}\kern-.125emX}}
\begin{document}

\title{
Spiderbot: An Open-Source Energy-Efficient Hexapod with Passive Gravity Compensation
}

\author{
  \begin{tabular}{c c c}
    Ritwik Sharma & Vimarsh Shah & Saransh Agrawal \\
    \small f20220470@goa.bits-pilani.ac.in & \small f20221060@goa.bits-pilani.ac.in & \small f20231123@goa.bits-pilani.ac.in \\
  \end{tabular}
  
  \\[0.6em]
  \small Electronics and Robotics Club, BITS Pilani, KK Birla Goa Campus
}

\maketitle

\begin{abstract}

Hexapod robots can achieve static stability with fewer actuated joints than bipeds or quadrupeds, yet many platforms still use 3-DOF legs, increasing weight and continuous torque requirement with limited gain in locomotion capability on flat, inclined and moderately rough terrains. We release Spiderbot, an open-source hexapod that uses a 4-bar linkage with a passive spring to mechanically support body weight, with a 2-DOF per-leg design that substantially reduces energy consumption. This mechanism substantially offloads gravitational torque during standing stance consuming only 1.5W (reduction of over 90\% over the unsprung version and up to 96\% over other similar hexapods). The passive spring compensation extends to payloads of up to 3.25kg with no additional torque requirements. The platform enables long-duration deployments on a modest battery budget and costs under \$400, making it suitable for large-scale multi-agent experiments. We validate the locomotion capabilities of the platform with an RL policy trained in mjlab, including successful sim-to-real transfer, despite the complexity of the mechanism. The platform is evaluated on flat and rough terrains, slope up to $15^\circ$ and step obstacles. We release all the CAD files, assembling instructions, and full training and deployment code along with the model checkpoints at \url{https://erc-bpgc.github.io/SpiderBot/}.
\end{abstract}

\begin{IEEEkeywords}
Actuation and Joint Mechanisms, Legged Robots, Model Learning for Control, Hexapods, Reinforcement Learning, Power Efficiency
\end{IEEEkeywords}

\section{\textbf{Introduction}}

Hexapod robots offer exceptional stability and versatility, making them ideal platforms for navigating unstructured and challenging environments. This stability, afforded by the redundant ground contacts of 6 limbs has driven their application in domains ranging from search and rescue and industrial inspection to planetary exploration \cite{10.3389/frobt.2024.1426269, six_link_mi13091404}. However, a fundamental trade-off persists in legged robot design: the balance between locomotive dexterity and energy efficiency. Conventional hexapods typically employ 3-DOF legs, but this complexity comes at a significant cost. The actuators in these systems must provide continuous torque simply to counteract gravity and maintain a standing posture, leading to high power consumption even during a standing stance \cite{power_hexa_7139915, Antbot}.

While numerous commercial and open-source platforms exist \cite{interbotix_hexapod, biomechanics_hexapod}, they often occupy two extremes: either they are low-cost but mechanically simple hobbyist kits with limited performance \cite{interbotix_hexapod, HexaV4}, or they are research platforms with costs reaching thousands of dollars \cite{HAntR, biomechanics_hexapod}. Furthermore, the high idle power draw of most conventional platforms across various scales \cite{arm2025efficientlearningbasedcontrollegged, lauronvisixleggedrobot, Antbot} makes long duration and large scale multi-agent deployment impractical without substantial infrastructure investment. There exists a clear need for a platform that bridges this gap, offering robust locomotion capabilities within an accessible, energy-efficient, and open-source framework.

\begin{figure}[!t]
    \centering
    \includegraphics[width=\columnwidth]{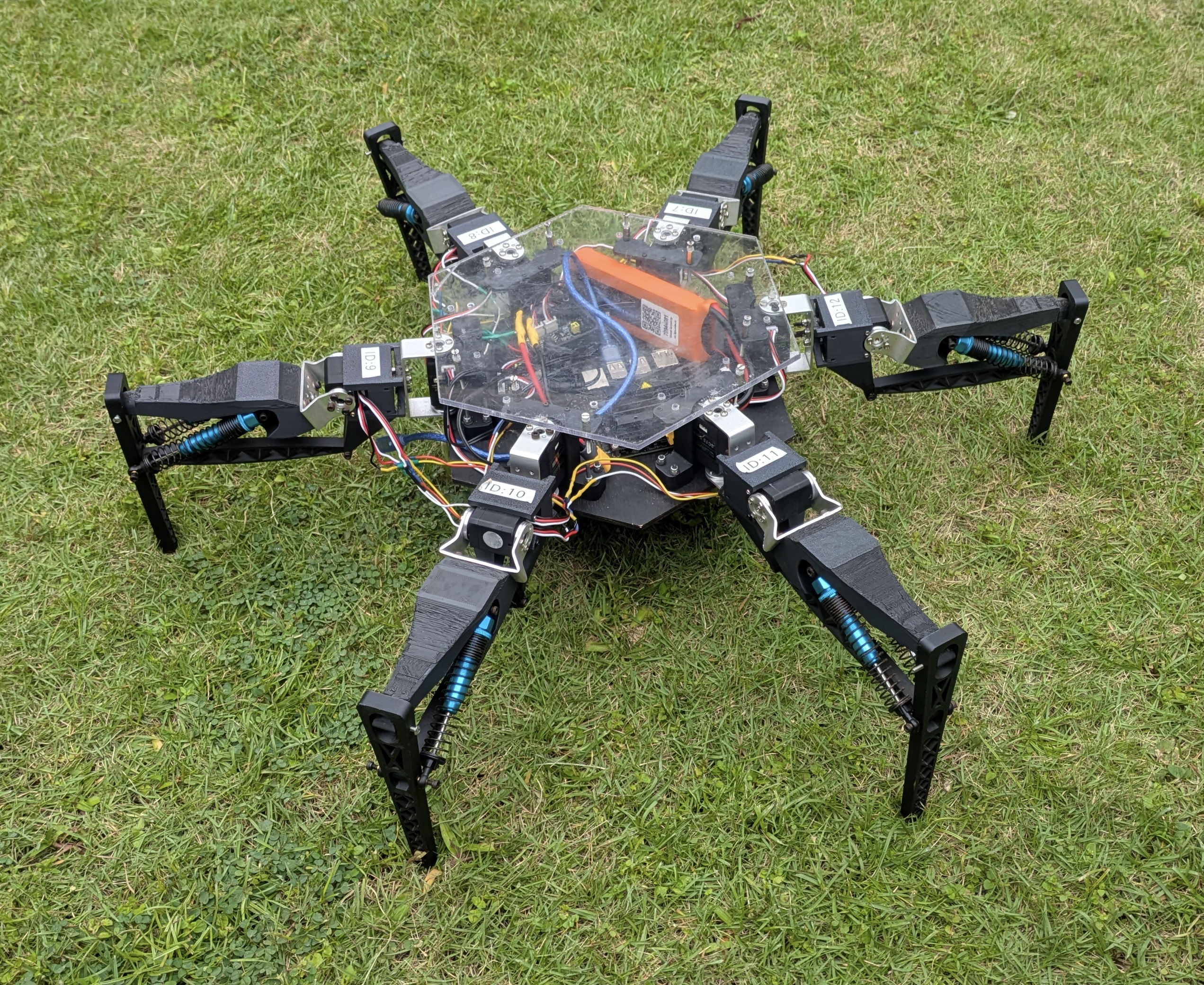}
    \caption{Spiderbot, the proposed open-source 12-DOF hexapod platform featuring spring-assisted four-bar legs for passive gravity compensation.}
    \label{fig:spidy_0}
\end{figure}

Our work makes the following contributions:
\begin{enumerate}
    \item An energy-efficient 2-DOF leg mechanism for hexapods that uses a passive spring-assisted four-bar linkage to achieve static weight support, reducing standing power consumption by \textbf{up to 96\%} compared to similar 3-DOF hexapods \cite{power_hexa_7139915, Antbot}. This advantage is sustained under payload, requiring almost no additional power up to a payload of \textbf{3.25 kg}.
    \item A quantitative analysis demonstrating improved energy efficiency in locomotion sustained under payload, with our design requiring \textbf{up to ~90\%} less power than the unsprung variant - at the cost of a maximum increase of about ~25\% in the cost of transport.
    \item The successful sim-to-real transfer of an RL-trained locomotion policy, despite the non-standard 4-bar linkage kinematics that cannot be represented in standard URDF format.
    \item The design and implementation of Spiderbot, a complete hexapod research platform (Fig.~\ref{fig:spidy_0}), manufacturable for under \$400 using 3D printing and off-the-shelf components, with the release of all hardware designs, training and deployment scripts as an open-source project.
\end{enumerate}

Our hexapod's mechanical design, code, and hardware details are completely open-source.

\section{\textbf{Related Works}}\label{sec:related}

Our research builds upon prior work in hexapod design, focusing on differentiating our approach in three key areas: mechanical architecture, cost-accessibility, and control methodology.

Most existing hexapod platforms, such as the popular PhantomX \cite{interbotix_hexapod} are based on a rigid, 18-DOF architecture (3-DOF per leg). While this configuration provides full control over foot placement, it is mechanically complex and energy inefficient \cite{power_hexa_7139915}. Bio-inspired designs have explored lightweight, compliant legs to improve efficiency \cite{biomechanics_hexapod}, but the integration of passive gravity-compensating mechanisms in low-cost platforms remains rare. Spiderbot’s 2-DOF design, which replaces the third actuated joint with a spring-loaded four-bar linkage, presents a new approach in this space, trading a DOF for substantial gains in power efficiency.

The cost of hexapod platforms is a significant factor for research and educational adoption. Commercial kits like the PhantomX MK-III retail for over \$1200 \cite{interbotix_hexapod}, while high-performance research robots such as HEBI's Daisy can cost upwards of \$80K \cite{biomechanics_hexapod}. Spiderbot, with a bill of materials under \$400, is positioned to fill the gap between low-performance hobby kits and expensive, closed-source commercial systems.

In terms of control, traditional hexapods rely on open-loop, inverse kinematics-based gait generators with hand-tuned parameters \cite{hexapod_terrain_control_2024}, which lack adaptability to terrain variation \cite{lee2020learning, HAntR}. Central Pattern Generators (CPGs) \cite{foundation_cpg_matsuoka1985, 
cmu_review_cpg_ijspeert2008central} offer smoother gait transitions but similarly require manual tuning and do not generalise well to unseen terrain \cite{yu2020enhancing, hwangbo2019learning}. Model-based planners address this by coupling proprioceptive feedback, foothold selection, and body pose optimisation for rough terrain negotiation \cite{HAntR, belter2016integrated}, at the cost of significant computational overhead. Recent work 
has shifted toward reinforcement learning, where policies trained in simulation with domain randomisation transfer robustly to real hardware \cite{tobin2017domain, hwangbo2019learning}, and hybrid CPG-RL methods achieve terrain-adaptive gaits while retaining interpretable gait structure \cite{terrain_adaptive_gait_2023, terrain_adaptive_cpg_rl_2023}. We adopt a learned velocity-tracking policy, as the constrained 2-DOF workspace of our mechanism makes hand-tuning a high-performing gait non-trivial, and we demonstrate successful sim-to-real transfer on our non-standard four-bar kinematics.

\section{\textbf{Mechanical Design}}\label{sec:mechanical}

\subsection{Hardware design requirements}

The mechanical design of Spiderbot was driven by a specific set of hardware requirements: (1) minimizing actuator energy expenditure during standing stance through passive weight support, (2) maintaining reliable locomotion performance over diverse locomotion tasks, including flat surfaces, rough surfaces and tasks like stair climbing and (3) low-cost and easily accessible manufacturing for components.
Passive stance support \cite{passivegravitycompensation} is achieved using a spring-linkage system that generates an elastic moment to compensate in part for the gravitational force on the body, hence reducing actuator efforts (Table~\ref{tab:payload}) during both static and dynamic tasks.

We employ a 2-DOF leg design based on a four-bar linkage, trading the 3 or more DOFs usually used in previous works \cite{interbotix_hexapod, phantomx_modeling_9624596} for energy-efficient operation while still achieving sustained smooth locomotion in both flat and moderately rough terrains. Note that the objective of this work is not to optimize the elastic, or structural parameters (link lengths, spring stiffness, masses etc.) but to introduce and analyze the proposed mechanism and establish its feasibility and practical benefits.

\begin{figure}[h!]
    \centering
    \includegraphics[width=1.0\linewidth]{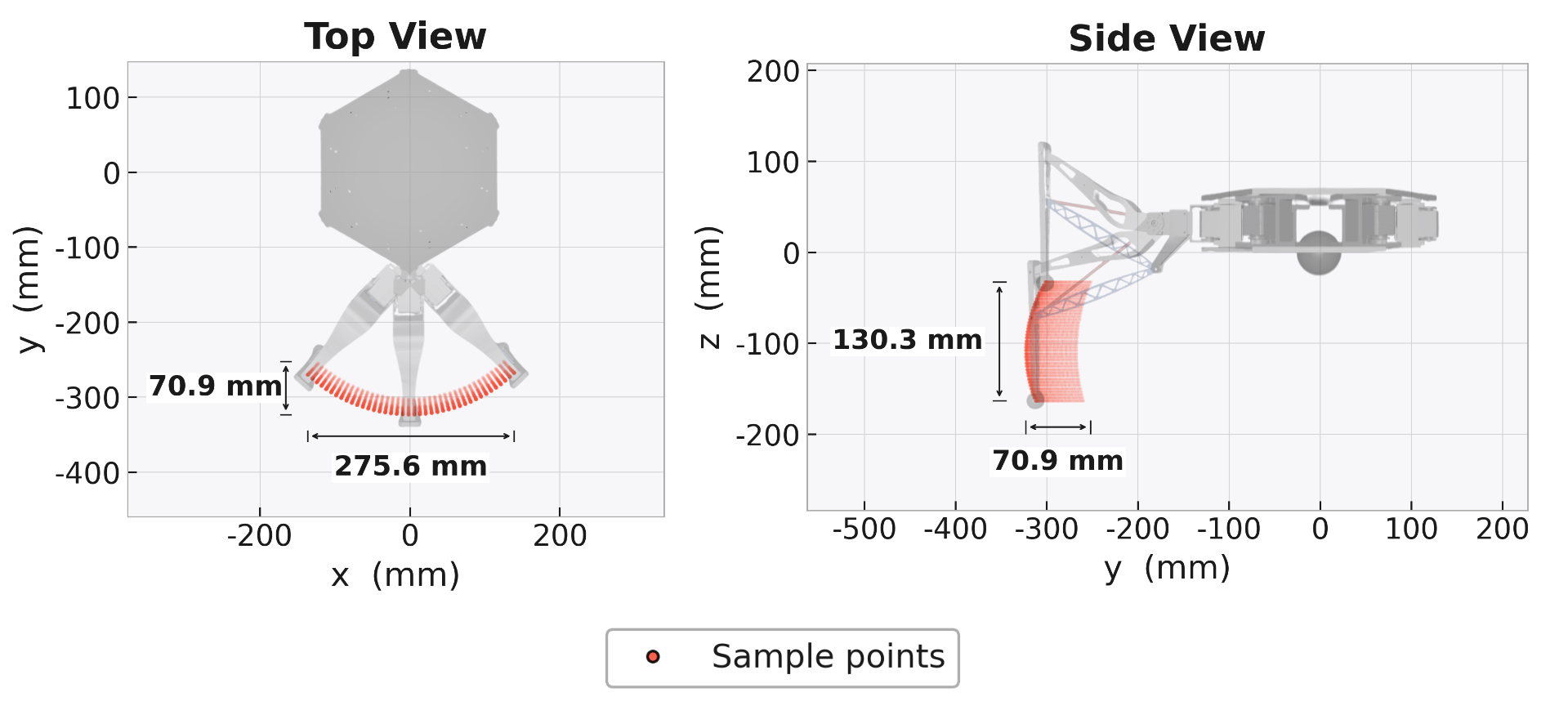}
    \caption{Foot workspace visualisation}
    \label{fig:foot_workspace}
\end{figure}

\begin{figure}[h!]
    \centering
    \includegraphics[width=1.0\linewidth]{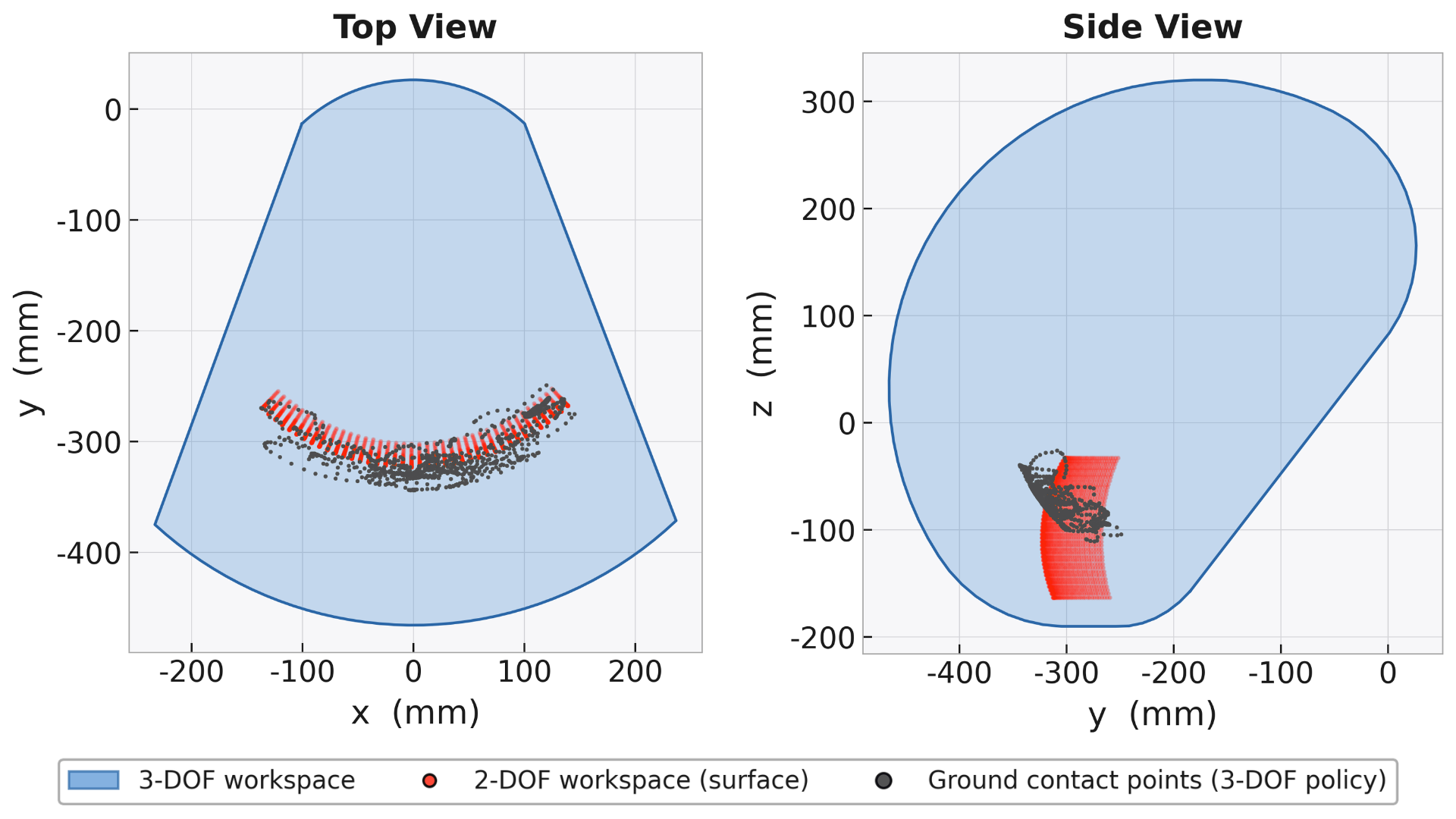}
    \caption{2-DOF vs 3-DOF leg workspace comparison}
    \label{fig:workspace_overlap}
\end{figure}

\subsection{Leg Design and Workspace}

Each leg of the robot is a 2-DOF mechanism based on a parallel 
four-bar linkage. The link lengths were determined through iterative 
kinematic analysis working back from two primary workspace 
requirements: a base ground clearance of at least \SI{10}{\centi\metre} 
during neutral stance, and a maximum foot lift height of \SI{5}{\centi\metre} 
above the ground contact point during dynamic locomotion. Starting 
from these constraints, forward kinematics were used iteratively to 
arrive at link lengths that simultaneously satisfy both requirements 
within the feasible range of the femur joint. The resulting workspace 
is visualised in Fig.~\ref{fig:foot_workspace}.

The geometry is designed to achieve passive gravity compensation \cite{passivegravitycompensation} during a standing stance. This is accomplished by embedding a spring mechanism that generates a restoring torque that effectively counteracts the gravitational load at the primary thigh (femur) joint. 

The 2-DOF nature of the design reduces the leg’s reachable workspace especially in the radial direction as compared to its 3-DOF counterpart. This is because the last link (tibia) is restricted to a vertical orientation at all times. This limits motions such as ducking under obstacles shorter than the height of the tibia link. To validate that the sufficiency of the 2-DOF workspace for dynamic locomotion, Fig. \ref{fig:workspace_overlap} overlays the reachable workspace of a 3-DOF leg configuration of our robot with that of the proposed 2-DOF design, along with ground-contact foot positions recorded during successful locomotion of the 3-DOF variant on rough terrain in simulation using a policy trained with an identical method as the one described in Section~\ref{sec:control}. The ground contact points are strongly 
concentrated in the region around the 2-DOF reachable workspace. This suggests that effective locomotion primarily exploits a subset of the full 3-DOF workspace that is well-covered by the 2-DOF 
mechanism, indicating that the reduction in kinematic degrees of freedom does not majorly sacrifice the workspace required for effective locomotion.

\vspace{10 pt}
\subsubsection{Notations}

The robot is divided into four free bodies: Link 1 (L1), Link 2 (L2), Link 3 (L3), and the Body (B).  
Reaction forces are expressed as $N_{XY}$, where the subscript $X$ indicates the interacting bodies and the subscript $Y$ indicates the direction (H = horizontal, V = vertical). For example, $N_{L1BH}$ represents the horizontal reaction force between L1 and B. Additional notation: $f_s$ denotes the spring force, $F$ the frictional force between L3 and the ground (or the legs and the ground), and $N$ the normal reaction between the legs and the ground.

\begin{figure}[htbp]
    \centering
    \includegraphics[width=1.0\linewidth]{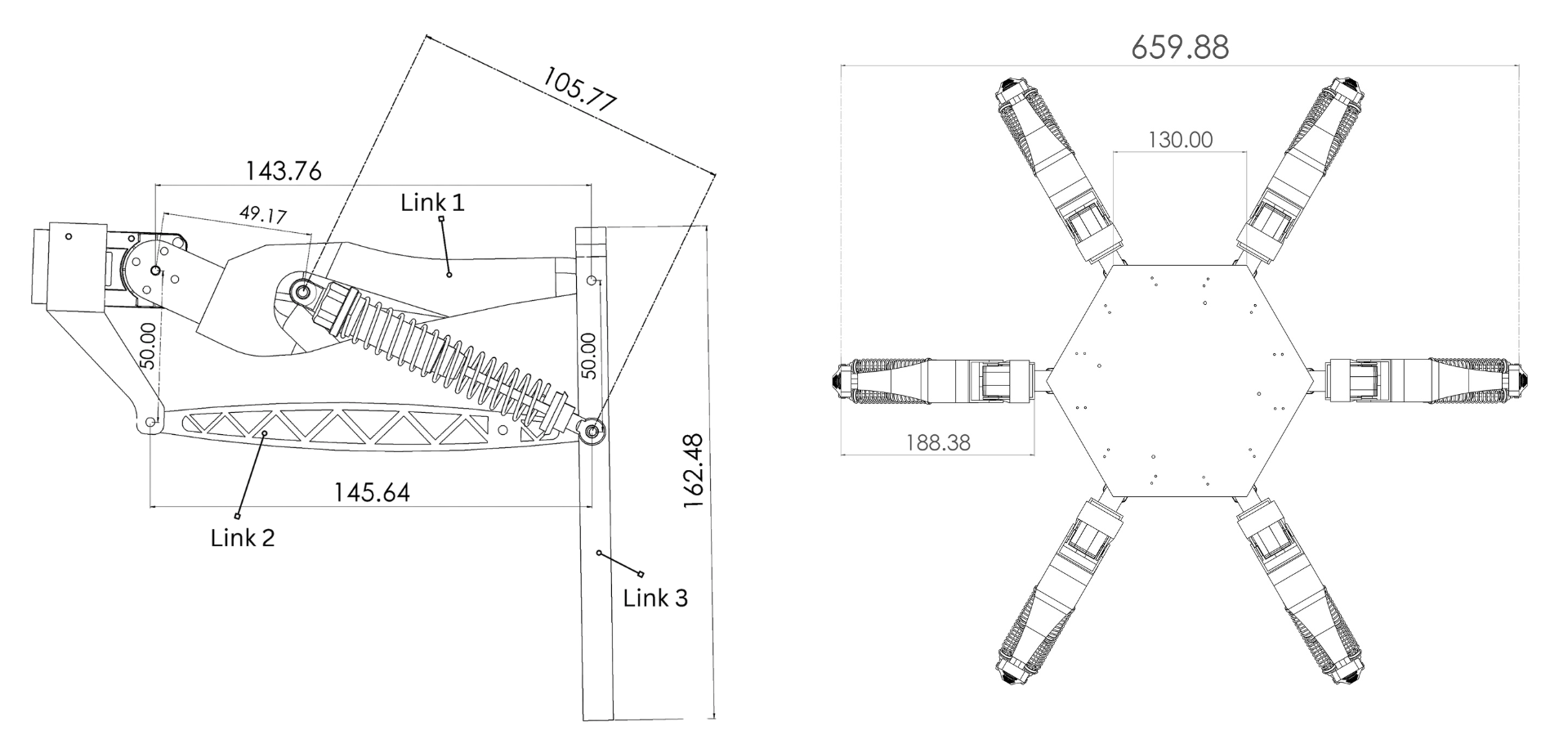}
    \caption{Leg design and dimensions for our hexapod (measurements in mm).}
    \label{fig:dimensions}
\end{figure}

\begin{figure}[htbp]
    \centering
    \includegraphics[width=0.55\linewidth]{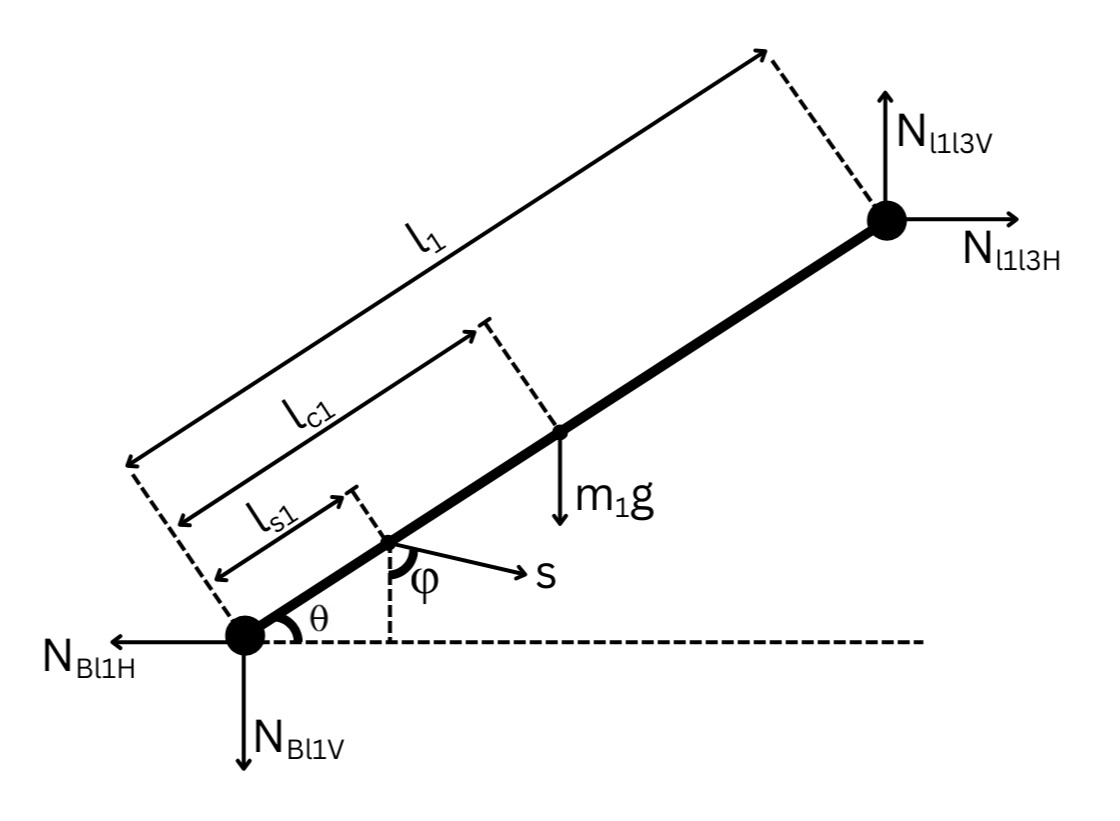}
    \caption{Free Body Diagram for link 1 (femur).}
    \label{fig:fbd_link1}
\end{figure}

\vspace{-10 pt}

\begin{table}[htbp]
\centering
\caption{Known physical quantities of the robot}
\begin{tabularx}{0.95\linewidth}{l l X}
\hline
\textbf{Symbol} & \textbf{Value} & \textbf{Description} \\
\hline
$m_1$ & $0.0608~\text{kg}$ & Mass of Link 1 (L1) \\
$m_2$ & $0.0125~\text{kg}$ & Mass of Link 2 (L2) \\
$m_3$ & $0.0124~\text{kg}$ & Mass of Link 3 (L3) \\
$m_{hb}$ & $2.59~\text{kg}$ & Mass of base Body (B) \\
$g$ & $9.81~m/s^2$ & Gravitational acceleration \\
$l_{L1}$ & $0.1463~\text{m}$ & Length of L1 \\
$l_{L3}$ & $0.1583~\text{m}$ & Length of L3 \\
$l_{CL1}$ & $0.0732~\text{m}$ & Distance of COM of L1 from pivot \\
$l_{CL2}$ & $0.0732~\text{m}$ & Distance of COM of L2 from pivot \\
$l_{CL3}$ & $0.0732~\text{m}$ & Distance of COM of L3 from top pivot \\
$l_V$ & $0.05~\text{m}$ & Vertical separation between L1 and L2 \\
$l_S$ & $0.05~\text{m}$ & Distance between top link pivot and spring pivot \\
\hline
\end{tabularx}
\end{table}

\vspace{10 pt}
\subsubsection{Kinematic and Static Analysis}
We conduct a static analysis for 2 stances: (i) A neutral standing stance with all six feet making contact with the ground. This is defined by link 1 making an angle of maximum 10 degrees with the horizontal, as it lies at the midpoint of the femur joint's actuation range, providing symmetrical motion capability for obstacle clearance (upward) and load absorption (downward).
(ii) A tripod stance with 3 alternate feet making contact with the ground and the other 3 feet at maximum retraction (30 degrees with horizontal). The tripod stance is chosen as it is the most energy efficient gait at higher speeds \cite{HexaV4} \cite{power_hexa_7139915}, and serves as a good representation of general locomotion costs.
The force diagram and equilibrium equations for a single link are expressed in Eq.~\ref{eq:1}, Eq.~\ref{eq:2}, Eq.~\ref{eq:3} and Fig.~\ref{fig:fbd_link1}.

\begin{figure}
    \centering
    \includegraphics[width=1.0\linewidth]{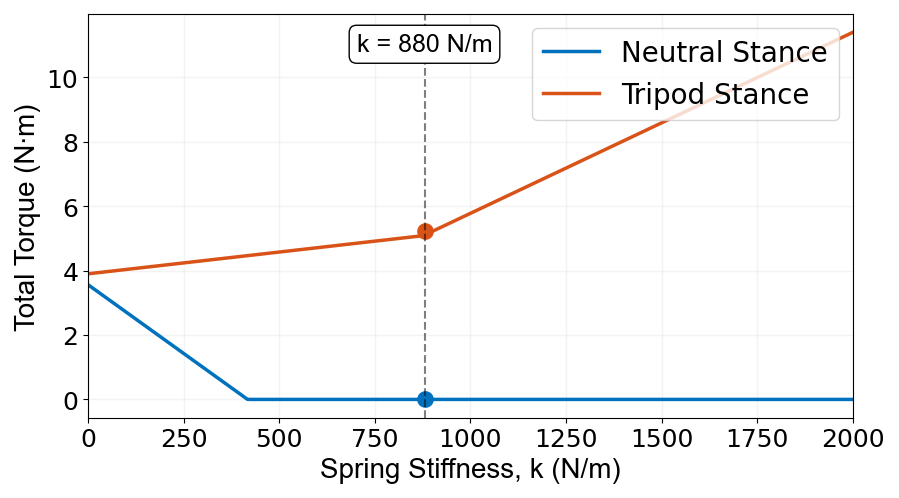}
    \caption{Total Robot Torque Requirement (Femur Joint) vs Spring Stiffness (k)}
    \label{fig:torque_vs_k}
\end{figure}

\vspace{-10pt}

{\small
\begin{equation} \label{eq:1}
-N_{HL1B} + f_s \sin \phi + N_{HLIL3} = 0  
\end{equation}
\begin{equation} \label{eq:2}
    N_{VL1L3} - N_{VL1B} - f_s \cos \phi = m_1g
\end{equation}
\begin{equation} \label{eq:3}
f_s l_S \cos(\phi - \theta) + N_{HL1L3} l_{L1} \sin \theta = - m_1 g \, l_{CL1} \cos \theta
\end{equation}
}
 \vspace{-6pt}

\begin{figure}
    \centering
    \includegraphics[width=0.8\linewidth]{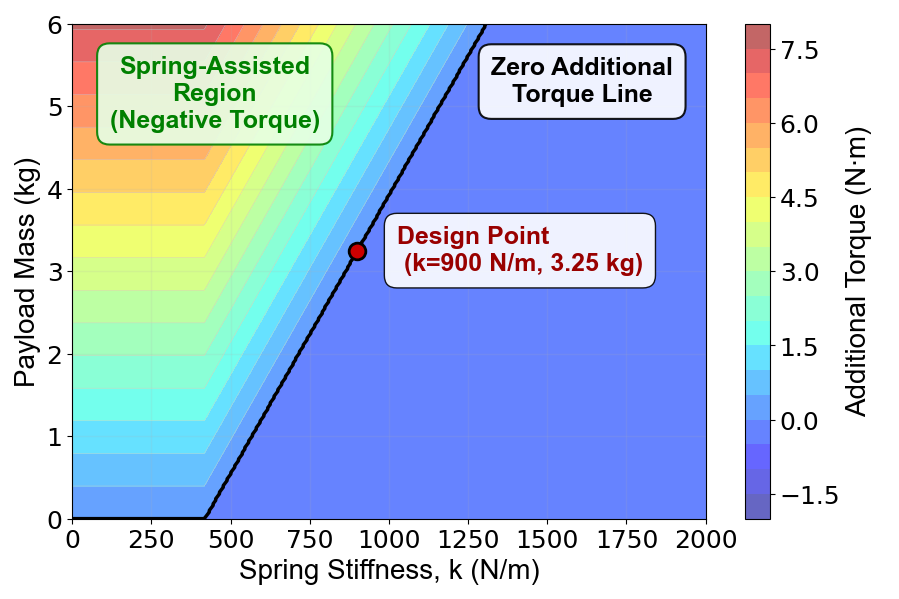}
    \caption{Additional torque requirement as a function of spring stiffness and payload mass for neutral stance configuration}
    \label{fig:additonal_torque_contour}
\end{figure}

The same formulation can be extended for all links of the robot, forming a complete system of equations. We plot the total femur joint torque requirement vs spring stiffness and the additional torque requirement for the neutral stance as a function of spring stiffness and payload capacity (Fig.~\ref{fig:torque_vs_k} and Fig.~\ref{fig:additonal_torque_contour}). Based on the inflection point of 880 N/m in the tripod stance torque curve (beyond which the locomotion cost rises steeply), we choose the spring constant as 900 N/m based on commercial availability (two 450 N/m springs per leg). A consequence of this choice is that it allows the robot to support payload of 3.25 kg with no additional torque. To calculate maximum motor torques we consider the tripod stance with a 3.25 kg payload. Since our actuator sizing is based on worst case static torques, an empirical safety margin of roughly 0.25 Nm is considered to manage unmodeled dynamics like inertial loading, contact forces. This is validated experimentally, with sustained experiments without actuator saturation. This yields the final required torque values of 10 kgcm for the hip (coxa) joint, and 20kgcm for the thigh (femur) joint.

\vspace{6px}
    
\begin{table}[!t]
\centering
\caption{Bill of Materials (BOM) for Spiderbot \label{tab:BOM}}
\label{tab:bom}
\renewcommand{\arraystretch}{1.1}
\begin{tabular}{@{}l c c@{}}
\toprule
\textbf{Component} & \textbf{Qty.} & \textbf{Total Cost (USD)} \\
\midrule
ST3215 High-Torque Serial Servo Motor & 6 & 126 \\
SC15 Compact Serial Servo Motor       & 6 & 108 \\
Arduino Uno Q Development Board       & 1 & 44 \\
PLA Filament Spool (1 kg)             & 1 & 10 \\
2-Cell 5200 mAh Li-Po Battery         & 1 & 30 \\
RC Shock Spring 450 N/m                & 12 & 18 \\
MPU-6050 6-DoF IMU Module             & 1 & 2  \\
Waveshare Serial-Servo Driver Board   & 2 & 20  \\
\midrule
\textbf{Total Estimated Cost}         &  -  & $\leq\textbf{400}$ \\
\bottomrule
\end{tabular}
\end{table}

\section{\textbf{System Electronics and Architecture}}\label{sec:electronics}

The electronic system uses commercial off-the-shelf components (as mentioned in Table~\ref{tab:bom}). An Arduino Uno Q handles high-level control and sensor processing and communicates with the motor drivers through a serial interface.

\begin{figure}[h!]
    \centering
    \includegraphics[width=\textwidth, height=6cm, keepaspectratio]{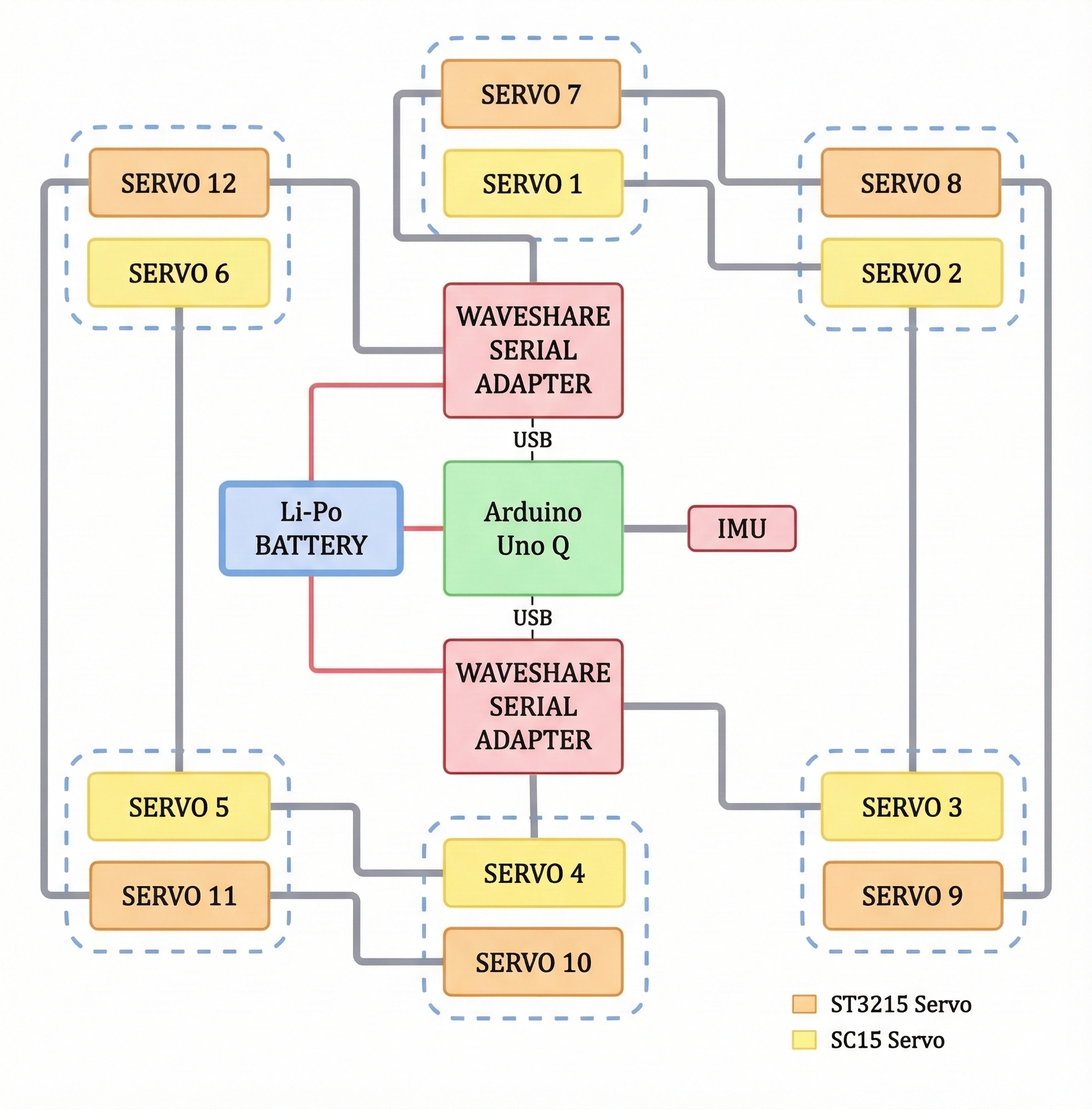}
    \caption{Electronics architecture of Spiderbot featuring compute, sensors and actuation.}
    \label{fig:robot_architecture}
\end{figure}

\subsection{Onboard Computation, Software, and Controls}
All onboard computation was performed by an Arduino Uno Q Dev Board, which executes the RL based control policy, processes IMU data for state estimation, and generates motor commands. An IMU connected via the GPIO header measures proprioceptive data, providing body orientation and angular velocity feedback. A Python code stack manages high-level control as well as communication with the motor drivers via serial communication.

The robot’s 12-DOF are actuated by serial bus servos that provide position feedback and reduce wiring. The servos chosen are the SC15 (15~kgcm at 7.4~V) for the coxa and the ST3215 (20~kgcm at 7.4~V) for the femur, based on the torque requirements derived in Section~\ref{sec:mechanical}. Both servos are commercially available off-the-shelf, consistent with the platform's low-cost design objectives. Two Waveshare Serial Bus Servo Driver boards control the servos, each interfacing with the Arduino through a dedicated USB connection. A 2-cell, 7.4 V, 5200 mAh LiPo battery powers the servo driver boards and the Arduino Uno Q.

\section{\textbf{Control}}\label{sec:control}

While classical controllers provide a strong analytical baseline, the 2-DOF four-bar leg limits foot placement and vertical workspace, and cannot vary step height independently of stride length, which reduces traction on uneven terrain and constrains its ability. These constraints motivate us to adopt a learning based control method that can adapt to these constraints and learn smooth locomotion on irregular ground profiles.

\subsubsection{Simulation Environment and Task Formulation}

The training environment is implemented using mjlab \cite{zakka2026mjlablightweightframeworkgpuaccelerated}, providing GPU-accelerated physics simulation through MJWarp integrated with Isaac lab's \cite{mittal2025isaaclab} manager based API. The simulation accurately models Spiderbot's mechanical constraints, including the 4-bar linkage geometry and passive spring dynamics. The model also incorporates MuJoCo's support for \textbf{tendons} to create the mechanical coupling in the 4-bar linkage, which cannot be defined by traditional URDF / XACRO format. The environment is configured to closely match hardware specifications for direct sim-to-real transfer. To improve sim-to-real transfer, zero-mean Gaussian noise is injected into the observation space as described in Table \ref{tab:domain_rand} emulating real sensor imperfections \cite{tobin2017domain}.

\begin{table}[ht]
\centering
\caption{Domain randomization parameters applied at training startup.}
\label{tab:domain_rand}
\resizebox{\columnwidth}{!}{%
\renewcommand{\arraystretch}{1.2}
\begin{tabular}{l l r}
\hline
\textbf{Parameter} & \textbf{Range} & \textbf{Operation} \\
\hline
Foot friction coefficient   & $[0.3,\ 1.2]$                  & Absolute \\
Encoder bias (rad)          & $[-0.015,\ 0.015]$             & Absolute \\
Base CoM offset (m)         & $[-0.025,\ 0.025]^{xy},\ [-0.03,\ 0.03]^{z}$ & Additive \\
Base link mass (kg)         & $[-1.0,\ 1.0]$                 & Additive \\
Limb mass scale             & $[0.8,\ 1.2]$                  & Scaled \\
Motor zero offset (rad)     & $[-0.035,\ 0.035]$             & Additive \\
PD gains ($k_p$, $k_d$)    & $[0.8,\ 1.2]$\textsuperscript{\dag}  & Scaled \\
Joint armature              & $[0.25,\ 2.0]$                 & Scaled \\
\hline
\multicolumn{3}{l}{\footnotesize\textsuperscript{\dag}Randomized per episode reset; all other terms randomized once at startup.}
\end{tabular}
}
\end{table}

The policy is trained in a procedurally generated environment composed of a $10 \times 20$ grid of terrain patches, each of size $8\,\text{m} \times 8\,\text{m}$, using a curriculum that progressively increases terrain difficulty as the agent improves \cite{rudin2022learningwalkminutesusing}. Seven terrain types are used during training,  as summarized in Table~\ref{tab:terrains}. Velocity commands are also introduced progressively via a staged curriculum, reaching final targets of $v_x, v_y \in [-0.5, 0.5]\,\text{m/s}$ and $\omega_z \in [-0.5, 0.5]\,\text{rad/s}$.

\begin{table}[ht]
\centering
\caption{Terrain types used during training.}
\label{tab:terrains}
\renewcommand{\arraystretch}{1.2}
\begin{tabularx}{\linewidth}{l c X}
\hline
\textbf{Terrain Type} & \textbf{Proportion} & \textbf{Parameters} \\
\hline
Flat            & 0.20 & Baseline flat ground. \\
Pyramid Stairs  & 0.20 & Step height: 2--5\,cm, step width: 0.5\,m. \\
Inverted Stairs & 0.20 & Step height: 2--5\,cm, step width: 0.5\,m. \\
Pyramid Slope   & 0.10 & Slope gradient: 0.05--0.35. \\
Inverted Slope  & 0.10 & Slope gradient: 0.05--0.35, inverted. \\
Random Rough    & 0.10 & Uniform noise height: 1.5--5\,cm. \\
Wave            & 0.10 & Sinusoidal surface, amplitude: 2.5--6.5\,cm. \\
\hline
\end{tabularx}
\end{table}

\subsubsection{State Representation}
The observation space is designed to provide sufficient information but try to have minimal sensor input for direct sim-to-real transfer. We use an \textbf{asymmetric actor-critic \cite{kumar2021rmarapidmotoradaptation}} setup, where the critic has access to privileged information during training, but the actor policy relies only on proprioceptive data to facilitate sim-to-real.

\begin{table}[ht]
\centering
\caption{Active reward terms and their weights.}
\label{tab:rewards}
\renewcommand{\arraystretch}{1.2}
\begin{tabular}{l l r}
\hline
\textbf{Category} & \textbf{Reward Term} & \textbf{Weight} \\
\hline
\multirow{2}{*}{Task}        & Track linear velocity  & $+5.00$ \\
                             & Track angular velocity & $+2.00$ \\
\hline
\multirow{2}{*}{Regulation}  & Upright orientation    & $+1.00$ \\
                             & Joint pose             & $+1.00$ \\
\hline
\multirow{6}{*}{Penalty}     & Joint position limits  & $-1.00$ \\
                             & Action rate (L2)       & $-0.10$ \\
                             & Foot clearance         & $-2.00$ \\
                             & Foot swing height      & $-0.25$ \\
                             & Foot slip              & $-0.10$ \\
                             & Soft landing           & $-1\times10^{-5}$ \\
\hline
\end{tabular}
\end{table}

\subsubsection{Policy Architecture and Reward Design}

The policy is trained with Proximal Policy Optimization (PPO) \cite{schulman2017proximalpolicyoptimizationalgorithms} using three-layer MLPs (512, 256, 128), for 5,000 iterations across 4096 parallel environments at 200 Hz with a control frequency of 50 Hz (decimation factor of 4).

\begin{figure*}[!th]
    \centering
    \includegraphics[width=0.95\textwidth]{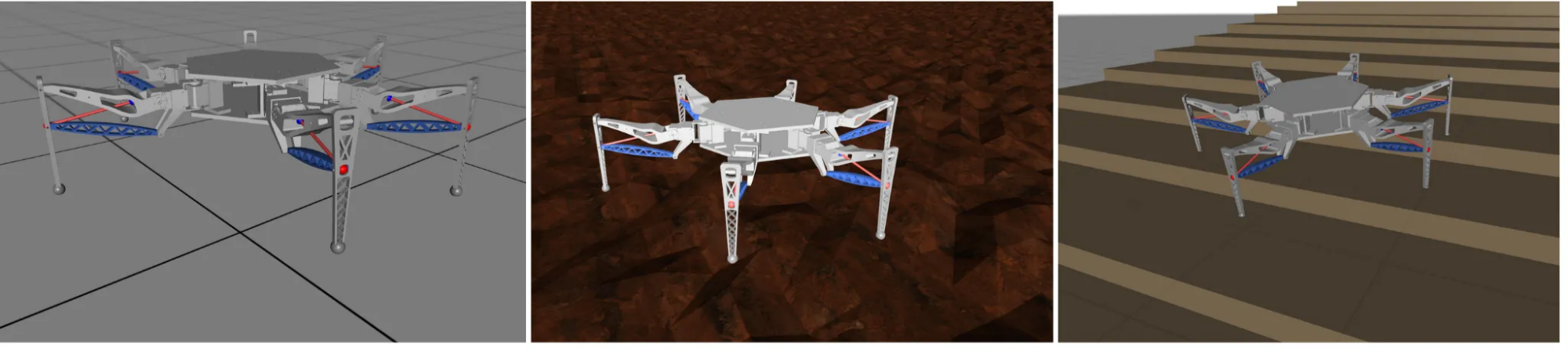}
    \caption{Different environments in MuJoCo simulation. From left to right: flat ground, rough terrain, stairs}
    \label{fig:mujoco_envs}
\end{figure*}

\section{\textbf{Experiments and Results}}
\label{sec:experiments}

We evaluate Spiderbot across a set of simulation
and hardware experiments designed to validate three primary claims:
(i) the 2-DOF leg design reduces overall power consumption compared
to conventional 3-DOF platforms, while the passive spring mechanism
further reduces static standing power at the cost of only a marginal
increase in locomotion CoT, with both advantages sustained under
payload, (ii) robust terrain traversal under varied environmental
conditions despite the constrained 2-DOF workspace, and (iii)
successful sim-to-real transfer of the learned locomotion policy on
the non-standard 4-bar leg kinematics.

\subsection{Experimental Setup}

All simulation experiments were conducted in MuJoCo at a control frequency of 50 Hz. Hardware experiments used the fully assembled 
platform (3.0 kg, 8V fixed supply, 2S 5200mAh LiPo). Base velocity was estimated using overhead camera from displacement over known duration. Each hardware condition was repeated N=10 trials; results report the mean.

\subsection{Locomotion on Flat Terrain}

Locomotion on flat ground is used as the primary baseline for all experiments. The robot was commanded to track velocity commands 
sampled from $v_x, v_y \in [-0.5, 0.5]$~m/s; a representative subset of commanded speeds is reported in Table~\ref{tab:flat}.

The Cost of Transport is defined as:
\begin{equation}
    \text{CoT} = \frac{P}{m g v}
\end{equation}
where $P$ is the mean power consumed (W), $m$ the robot mass (kg), $g = \SI{9.81}{\metre\per\second\squared}$, and $v$ the mean forward velocity (m/s). In simulation, $P = \frac{1}{T}\sum_i \sum_t |\tau_i \, \dot{\theta}_i| \, \Delta t$ over all joints $i$ and timesteps. On hardware, $P = \frac{1}{T} V \int I \, \mathrm{d}t$. The CoT and average speed are reported in Table~\ref{tab:flat}. The simulation CoT at maximum commanded speed is competitive with HAntR's reported tripod values, while hardware CoT is higher across all conditions. The robot achieves approximately 85\% of the commanded velocity on hardware at the lowest command, with tracking degrading at higher speeds.

\begin{table}[h]
\centering
\caption{Flat Terrain Locomotion Performance (mean, $N=10$)}
\label{tab:flat}
\begin{tabular}{lcccccc}
\hline
\textbf{Speed} & \multicolumn{2}{c}{\textbf{CoT}} & \multicolumn{2}{c}{\textbf{Avg Speed (m/s)}} \\
\textbf{(m/s)} & Sim & Real & Sim & Real\\
\hline
0.2 & 1.936  & 6.22  & 0.192  & 0.163\\
0.35 & 1.003  & 5.034  & 0.373  & 0.186\\
0.5 & 0.960  & 3.54  & 0.4305  & 0.231\\
\hline
\end{tabular}
\end{table}

\subsection{Effect of Payload on Power and Locomotion}

To validate the analytical prediction that the spring mechanism sustains its energy advantage under load, the robot was tested 
carrying different payloads, distributed symmetrically on the body frame. For each payload condition, static standing power consumption and CoT at a fixed commanded velocity of $0.2$m/s are recorded. To isolate the contribution of the spring each condition is also tested with the spring removed, providing a direct sprung vs.\ unsprung comparison across payload levels. The static standing power results (Table~\ref{tab:payload}) confirm the analytical prediction from Section~\ref{sec:mechanical}: the passive spring mechanism reduces standing power consumption by over 90\% compared to the unsprung configuration, with this advantage transferring under payload. Notably, the 3.25\,kg payload condition incurs no additional static torque requirement, validating the spring stiffness selection. The increase of $\sim$25\% in locomotion CoT introduced by the spring, represents a favourable trade-off for applications prioritising endurance and payload-carrying over peak locomotion efficiency. The maximum payload tested while maintaining a structured gait and a velocity of at least 0.05 m/s was 3.75 kg.

\begin{table}[h]
\centering
\caption{Effect of Payload on Standing Power and CoT (sprung vs.\ unsprung, hardware)}
\label{tab:payload}
\begin{tabular}{lcccc}
\hline
\textbf{Payload (kg)} & \multicolumn{2}{c}{\textbf{Standing Power (W)}} 
& \multicolumn{2}{c}{\textbf{CoT at 0.2 m/s (commanded)}} \\
 & Sprung & Unsprung & Sprung & Unsprung \\
\hline
0.0 & 1.5 & 15 & 6.22 & 4.97 \\
1.25 & 1.57 & 21.21 & 8.41 & 6.79\\
2.5 & 1.65 & 27.80 & 15.65 & 12.71 \\
3.25 & 1.76 & 33.92 & 27.63 & 23.10 \\
3.75 & 4.11 & 39.87 & 49.13 & 40.74 \\
\hline
\end{tabular}
\end{table}

\begin{figure}[htbp]
    \centering
    \includegraphics[width=0.95\linewidth]{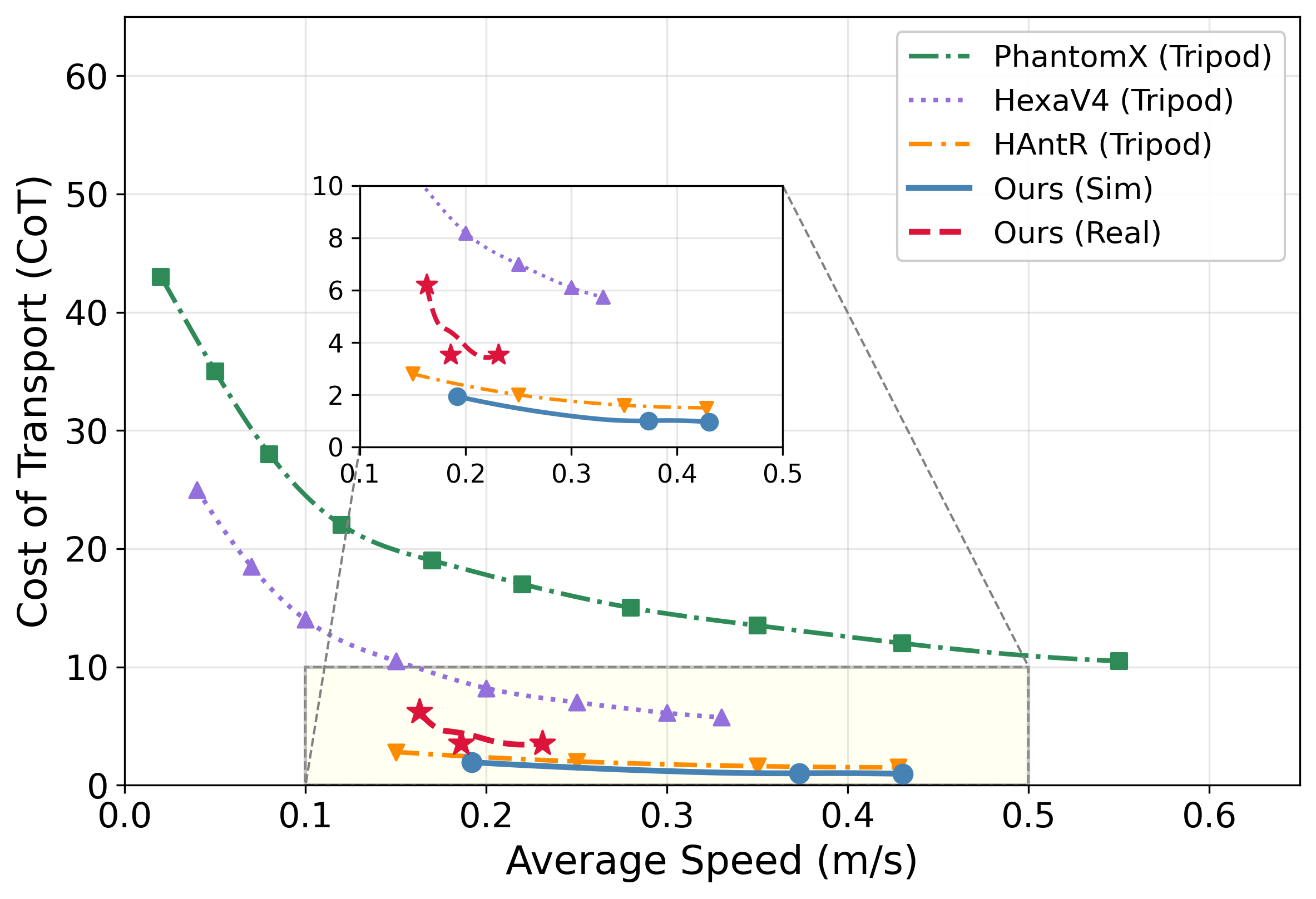}
    \caption{CoT vs speed comparison.}
    \label{fig:cot_vs_speed}
\end{figure}

\begin{figure}[htbp]
    \centering
    \includegraphics[width=0.9\linewidth]{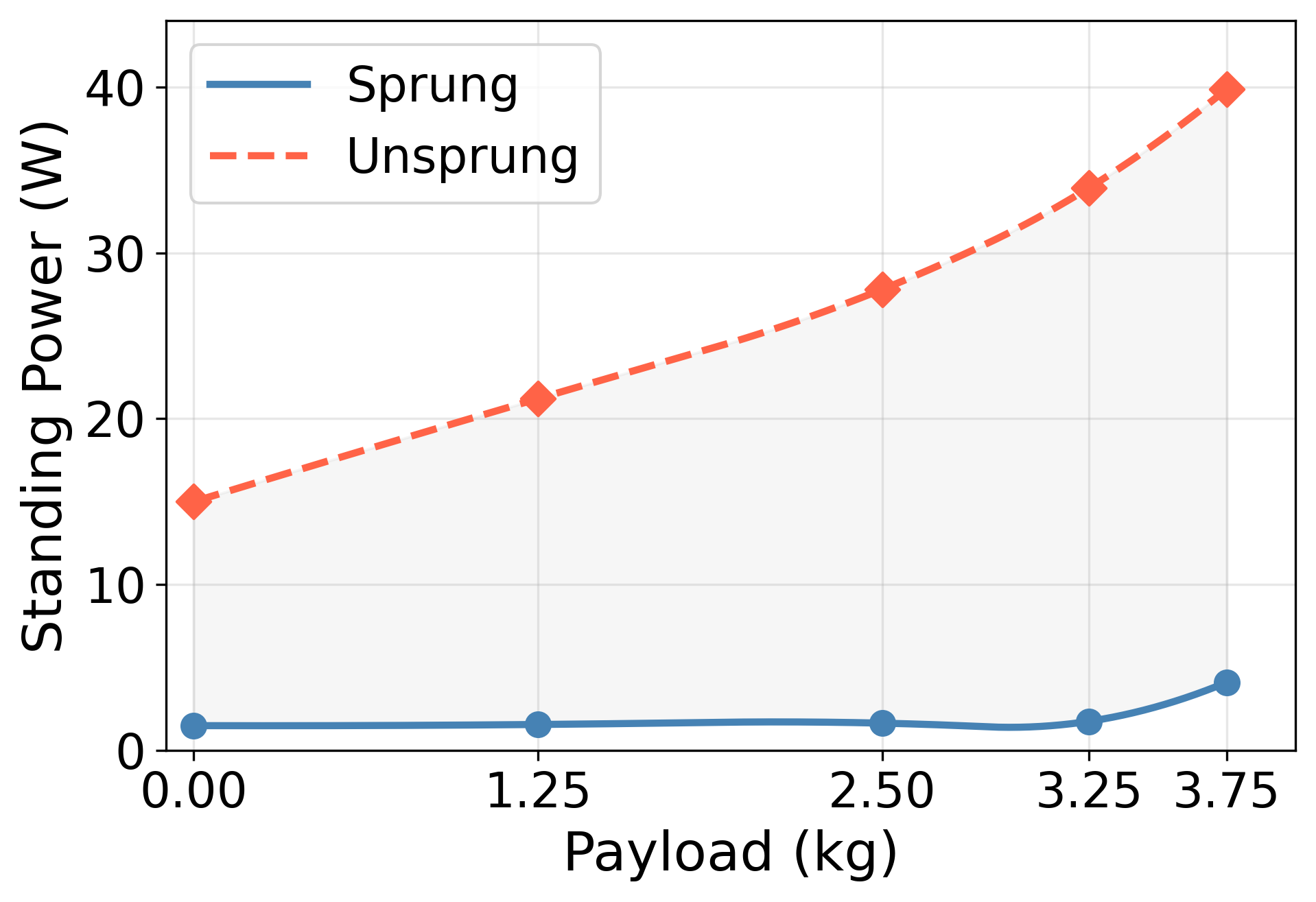}
    \caption{Standing Power vs Payload.}
    \label{fig:standing_power_vs_payload}
\end{figure}

\subsection{Terrain Traversal}

To evaluate locomotion robustness on inclined surfaces, the robot's ascent was tested on slopes with different angles. A trial is marked successful if the robot traverses \SI{2}{\metre} along the slope without falling or requiring manual intervention. The robot successfully ascends slopes up to 15° in both sim and hardware, and the recorded CoT and average speed are reported in Table~\ref{tab:slope}. Discrete stair traversal evaluates the platform's ability to negotiate step-type discontinuities. Obstacle crossing succeeds up to 60\% of maximum foot lift in simulation and 40\% on hardware (Table~\ref{tab:obstacle}). The 100\% case fails in both sim and hardware, representing a hard kinematic limit of the 2-DOF design: the fixed tibia orientation prevents the foot from clearing obstacles at or beyond the maximum lift height regardless of policy behaviour. To test generalisation beyond structured obstacles, the robot was evaluated on randomly generated rough terrain. In simulation, terrain height fields were implemented as a heightfield mesh generated from a greyscale PNG image, spanning a 10×10 m area with a maximum surface elevation of 5 cm. On hardware, a \SI{15}{\metre} patch spanning two surface types: unpacked natural soil and an uneven paved surface with irregular tile height variation was used. Rough terrain CoT increases by 2.1× over flat in simulation and 2.8× on hardware (Table~\ref{tab:rough}).

\begin{table}[h]
\centering
\caption{Slope Traversal Results}
\label{tab:slope}
\begin{tabular}{lcccccc}
\hline
 & \multicolumn{2}{c}{\textbf{Success (\cmark/\xmark)}} & \multicolumn{2}{c}{\textbf{CoT}} & \multicolumn{2}{c}{\textbf{Avg Speed (m/s)}} \\
\textbf{Condition} & Sim & Real & Sim & Real & Sim & Real \\
\hline
$5^\circ$ Ascent  & \cmark & \cmark & 1.48 & 8.28 & 0.225 & 0.12 \\

$10^\circ$ Ascent  & \cmark & \cmark & 2.27 & 11.14 & 0.15 & 0.09 \\

$15^\circ$ Ascent  & \cmark & \cmark & 2.46 & 9.54 & 0.12 & 0.12 \\
\hline
\end{tabular}
\end{table}

\begin{table}[h]
\centering
\caption{Obstacle Crossing Results}
\label{tab:obstacle}
\begin{tabular}{lccccc}
\hline
\textbf{Obstacle} & \textbf{Height/Leg} & \multicolumn{2}{c}{\textbf{Success (\cmark / \xmark)}} & \multicolumn{2}{c}{\textbf{CoT}} \\
\textbf{Height (cm)} & \textbf{Length (\%)} & Sim & Real & Sim & Real \\
\hline
2.0 & 40\% & \cmark & \cmark & 1.17 & 12.69 \\
3.0 & 60\% & \cmark & \xmark & 1.84 & - \\
5.0 & 100\% & \xmark & \xmark & - & - \\
\hline
\end{tabular}
\end{table}

\begin{table}[!h]
\centering
\caption{Rough Terrain Performance (mean, $N=10$)}
\label{tab:rough}
\begin{tabular}{lcccc}
\hline
 & \multicolumn{2}{c}{\textbf{CoT}}\\
\textbf{Terrain} & Sim & Real\\
\hline
Flat (baseline) & 0.96 & 3.54 \\
Rough terrain   & 2.03 & 9.79 \\
\hline
\end{tabular}
\end{table}

\subsection{Sim-to-Real Transfer Analysis}

An attempt is made to quantify the robustness of the sim-to-real transfer across terrain types. For each terrain condition, CoT and nominal speed are reported in both simulation and on hardware under matched commanded velocities (Table~\ref{tab:s2r}). The gap in CoT is substantially larger than the corresponding gap in speed across all terrain types, with speed transferring to within approximately 50\% of simulation values while CoT gaps exceed 3$\times$ on flat terrain and 4$\times$ on slopes.

\begin{table}[!h]
\centering
\caption{Sim-to-Real Transfer Gap}
\label{tab:s2r}
\begin{tabular}{llccc}
\hline
\textbf{Terrain} & \textbf{Metric} & \textbf{Sim} & \textbf{Real} \\ 
\hline
Flat       & CoT              & 0.96 & 3.54 \\ 
Flat       & Speed (m/s)      & 0.43 & 0.23 \\ 
$10^\circ$ Slope & CoT         & 2.27 & 11.14 \\ 
$10^\circ$ Slope & Speed (m/s) & 0.15 & 0.09 \\ 
Rough      & CoT              & 2.03 & 9.79 \\ 
Rough      & Speed (m/s)      & 0.289 & 0.11 \\ 
\hline
\end{tabular}
\end{table}

\subsection{Comparative Benchmarking}

To contextualise the platform's energy efficiency, both static standing power and steady-state CoT on flat terrain at nominal walking speed are compared against existing hexapod platforms (Table~\ref{tab:benchmark}, Fig.~\ref{fig:cot_vs_speed}). Standing power is highlighted separately as it directly reflects the benefit of passive gravity compensation independent of gait or control strategy, enabling a fair mechanism-level comparison across platforms. Where direct data is unavailable, values are taken from the respective published papers under the closest matching speed and load condition. Our platform consumes an order of magnitude less standing power than the PhantomX and approximately 6$\times$ less than HexaV4, with the gap attributable to the combined effect of eliminating one actuated joint per leg and passive spring offloading of gravitational torque. Note that PhantomX and HexaV4 CoT values used in Fig.~\ref{fig:cot_vs_speed} are inferred from data points in their respective papers, as tabulated values were not directly available.

\begin{table}[t]
\caption{Comparison with Existing Hexapod Platforms}
\label{tab:benchmark}
\centering
\begin{tabular}{lccc}
\toprule
\textbf{Platform} & \textbf{Stand. Power (W)} & \textbf{DOF/leg} \\
\midrule
Spiderbot                   & 1.5                    & 2 + spring \\
HAntR \cite{HAntR}     & -$^\dagger$    & 4 \\
PhantomX \cite{interbotix_hexapod} & $\sim$40         & 3 \\
HexaV4 \cite{HexaV4} & $\sim$10            & 3 \\
\bottomrule
\multicolumn{3}{p{0.9\linewidth}}{\footnotesize $^\dagger$Standing power not reported \cite{HAntR}.}\\
\end{tabular}
\end{table}

\subsection{Discussion}
The results collectively validate the core design trade-off: the passive spring mechanism delivers substantial static efficiency gains that persist under payload, at the cost of a modest locomotion CoT increase and a constrained kinematic workspace. The standing power reduction is attributable to two compounding factors, the elimination of one actuated joint per leg and the spring offloading gravitational load passively. The unsprung configuration expectedly exhibits lower locomotion CoT than the sprung case at matched payloads, since the fixed restoring force of the spring introduces a resistive torque that the policy must work against dynamically. The workspace constraint manifests predictably as a hard ceiling on negotiable obstacle height, but does not prevent generalisation to slopes or rough terrain, suggesting the RL policy compensates through learned timing rather than explicit foot placement for total body reorientation. The mechanism thus deliberately trades obstacle-negotiation dexterity for passive support, lower standing power, simplicity, and cost. The fixed spring stiffness provides diminishing benefit beyond the 3.25~kg design point and cannot be adapted to different terrain or loading conditions without physical hardware changes, representing the primary mechanical limitation of the current design. The sim-to-real gap in CoT is substantially larger than the corresponding gap in speed across all terrain conditions, which is consistent with a systematic difference between mechanical joint power computed in simulation and total electrical draw on hardware. This includes driver losses, communication overhead, back-EMF effects and low-cost hardware not matching manufacturer ratings, the direct effect of which is a discrepancy observed in other servo-driven platforms~\cite{Antbot} too. Actuator/electrical modelling would be required for accurate power prediction. Together, these results suggest the platform is well-suited to endurance and payload-focused deployments where static efficiency and accessibility outweigh the need for maximum kinematic dexterity.

\section{\textbf{Conclusion}} \label{sec:conclusion}

We present Spiderbot, a low-cost, energy-efficient hexapod platform featuring a spring-assisted 2-DOF four-bar leg mechanism. Passive gravity compensation reduces standing power to 1.5 W, over 90$\%$ lower than the unsprung configuration, while supporting payloads up to 3.25 kg with no additional actuator torque requirement.

The RL-trained policy achieves locomotion on flat ground, slopes up to 15\textdegree, and step obstacles up to 40$\%$ of the maximum foot lift height, with successful sim-to-real transfer despite the non-standard four-bar kinematics. The passive spring introduces an approximately 25$\%$ increase in locomotion CoT, representing a trade-off against the substantial reduction in standing power.

At a total cost of under \$400 with fully open-sourced hardware and software, Spiderbot provides an accessible platform for legged robotics research, particularly for endurance- and payload-focused applications where static efficiency is prioritized over maximum kinematic dexterity.

Future work will focus on improving actuator and electrical modeling to reduce the sim-to-real energy gap, while exploring adaptable compliance and higher-dexterity leg designs that retain Spiderbot’s low standing power and accessibility.

\section*{Acknowledgment}

We would like to thank Akshat Tubki, Adarsh Gopalakrishnan, Aadya Keni, Aasim Sayyed, Ansh Parmeshwar, Kshitij Takale, Garv Gupta and other members of Electronics and Robotics Club, BITS Pilani - Goa for their contributions to the development of the project.

We would also like to thank BITS Pilani, KK Birla Goa Campus and Sandbox Innovations Lab for their support in terms of equipment use and funding for the project.


\renewcommand{\bibfont}{\footnotesize}
\printbibliography

\vspace{12pt}
\end{document}